\documentclass[conference]{IEEEtran}
\usepackage{times}
\usepackage{graphicx} 
\usepackage{booktabs} 
\usepackage{caption}
\usepackage{multirow}
\usepackage{makecell}
\usepackage{xcolor}    
\newcommand{\cmark}{\textcolor{green}{\checkmark}}
\newcommand{\xmark}{\textcolor{red}{$\times$}}
\usepackage{amsmath} 
\usepackage{amssymb} 
\usepackage{caption}
\usepackage[numbers]{natbib}
\usepackage{multicol}
\usepackage[bookmarks=true]{hyperref}

\begin{document}

\title{Constructing the Umwelt: Cognitive Planning through Belief-Intent Co-Evolution}

\author{\authorblockN{Shiyao Sang}
\authorblockA{Independent Researcher, Jiangsu, China \\
Email: www.fengmao@outlook.com}}

\maketitle

\begin{abstract}
This paper challenges a prevailing epistemological assumption in End-to-End Autonomous Driving: that high-performance planning necessitates high-fidelity world reconstruction. Inspired by cognitive science, we propose the Mental Bayesian Causal World Model (MBCWM) and instantiate it as the Tokenized Intent World Model (TIWM), a novel cognitive computing architecture. Its core philosophy posits that intelligence emerges not from pixel-level objective fidelity, but from the Cognitive Consistency between the agent's internal intentional world and physical reality.
By synthesizing von Uexküll's \textit{Umwelt} theory, the neural assembly hypothesis, and the triple causal model (integrating symbolic deduction, probabilistic induction, and force dynamics) into an end-to-end embodied planning system, we demonstrate the feasibility of this paradigm on the nuPlan benchmark. Experimental results in open-loop validation confirm that our Belief-Intent Co-Evolution mechanism effectively enhances planning performance. Crucially, in closed-loop simulations, the system exhibits emergent human-like cognitive behaviors, including map affordance understanding, free exploration, and self-recovery strategies.
We identify Cognitive Consistency as the core learning mechanism: during long-term training, belief (state understanding) and intent (future prediction) spontaneously form a self-organizing equilibrium through implicit computational replay, achieving semantic alignment between internal representations and physical world affordances. Based on this, we propose two fundamental hypotheses: (H1) The efficacy of embodied intelligence depends less on parameter scale or data volume, and more on the degree of semantic alignment between internal attentional dynamics and physical affordances; (H2) The Intent Token serves as a functional neural unit, computationally analogous to the biological neural assembly. TIWM offers a neuro-symbolic, cognition-first alternative to reconstruction-based planners, establishing a new direction: planning as active understanding, not passive reaction.
\end{abstract}

\IEEEpeerreviewmaketitle

\section{Introduction}
End-to-End Autonomous Driving aims to map raw sensory data directly to control actions, thereby transcending the modular decoupling inherent in traditional ``Perception-Prediction-Planning" pipelines~\cite{huPlanningOrientedAutonomousDriving2023a, jiangVADVectorizedScene2023b}. Recent frameworks, such as Sparse Scene Representation (SSR)~\cite{liNavigationGuidedSparseScene2024}, have demonstrated that superior performance can be achieved without compromising efficiency through the introduction of sparse Bird’s Eye View (BEV) representations and future BEV self-supervision. However, the optimization of these systems remains fundamentally dependent on dense reconstruction supervision of the future~\cite{liNavigationGuidedSparseScene2024}. Furthermore, their internal representations frequently lack interpretable cognitive semantic structures, which fundamentally limits their capacity to support long-term, consistent intent evolution.

\begin{figure}[ht]
\vspace{5pt}
\centering
\begin{minipage}[b]{1\linewidth}
\includegraphics[width=\linewidth]{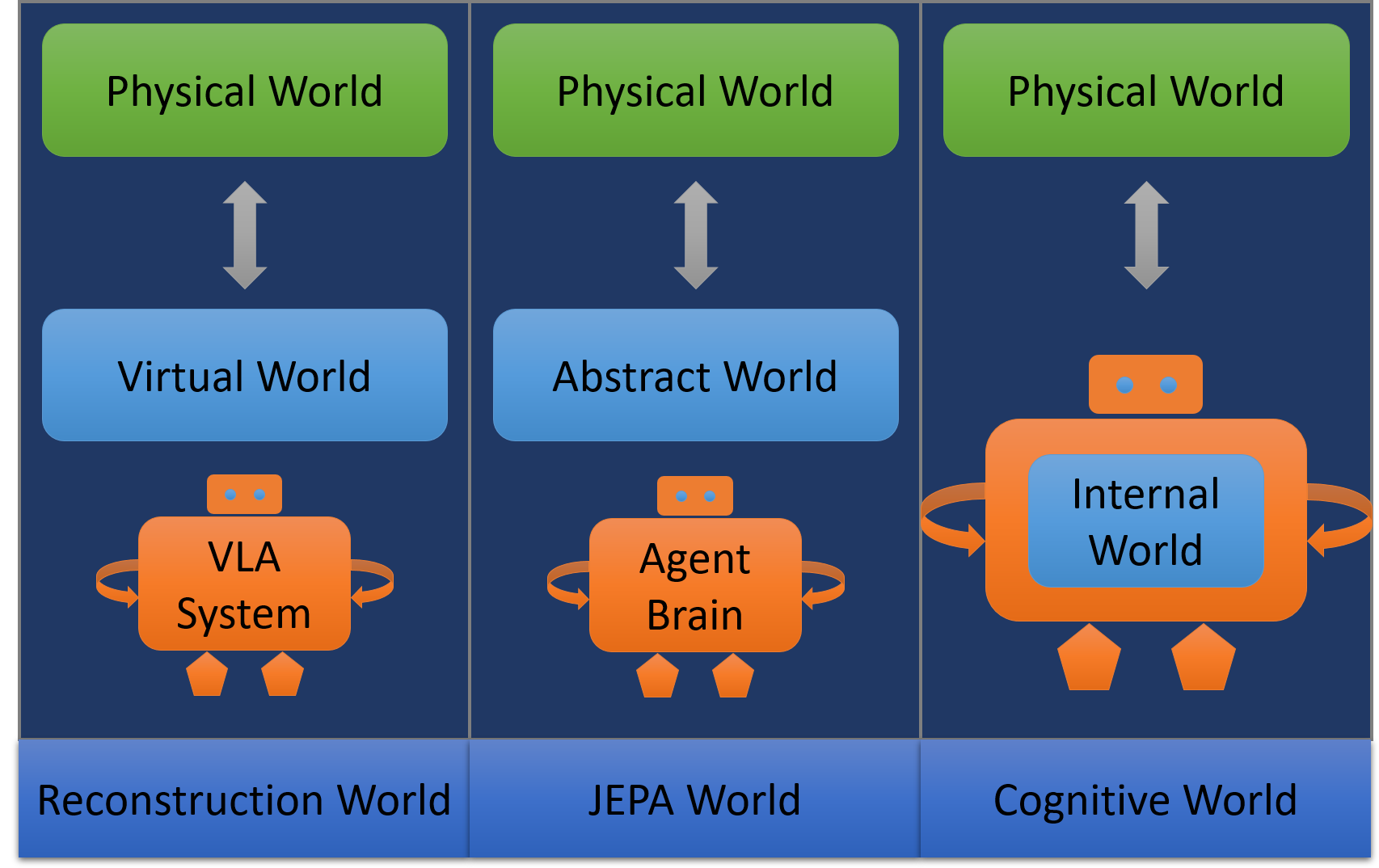}
\caption{\small \textbf{Comparison of Distinct World Model Paradigms.} (Left) Reconstruction-based World Models attempt to replicate the physical world by training dense models to reconstruct visual inputs. (Middle) LeCun’s Joint-Embedding Predictive Architecture (JEPA)~\cite{lecunPathAutonomousMachine2022} aims to extract abstract laws of the physical world, serving as a modular component attached to the agent's brain. (Right) Our proposed MBCWM constructs an Internal Cognitive World. Through a Belief-Intent Co-Evolution mechanism, it achieves alignment between the robot’s internal cognitive state and the affordances of the physical world.}
\label{fig:worldmodel}
\end{minipage}
\vspace{-35pt}
\end{figure}

At a deeper level, current mainstream methods share a deeply rooted philosophical assumption: high-performance planning must be built upon a dense reconstruction of future states—that is, via a high-fidelity world model combined with an end-to-end architecture characterized by dense internal representations (Fig.~\ref{fig:worldmodel}). While effective for short-term, local tasks, this paradigm inevitably leads to two fundamental limitations: (1) Computational Redundancy: Models are forced to model vast amounts of task-irrelevant environmental details. (2) Cognitive Myopia: Due to the lack of explicit long-term belief maintenance mechanisms, the planning process remains constrained by short-sighted Markovian assumptions, failing in complex scenarios requiring cross-temporal causal reasoning.

We argue that this assumption is fundamentally flawed. Synthesizing insights from cognitive science, biology, and computational theory, we identify three major limitations of the current paradigm: First, Reconstruction is Redundancy. Reconstructing the world at the pixel level is akin to brute-force decryption of reality---it introduces noise, computational waste, and semantic ambiguity~\cite{lecunPathAutonomousMachine2022,heMaskedAutoencodersAre2022}. Biological intelligence---from insects to humans---never reconstructs the world; it extracts task-relevant invariants~\cite{uexkuellUmweltUndInnenwelt1909}. Within a computational system, planning and world modeling share a unified generative ``Seed''; consequently, explicit reconstruction is not only superfluous but mathematically redundant. Instead, our approach adopts a ``Side-Channel Attack''~\cite{kocherTimingAttacksImplementations1996} strategy, extracting only the ``Private Key'' required for planning--Intent--via sparse semantic tokens. Second, Language is not Embodiment. Large Language Models (LLMs) are essentially containers of collective human memory but lack proprioception—the sensory-motor closed loop of interacting with the physical world in real-time~\cite{tianDriveVLMConvergenceAutonomous2025, zhouOpenDriveVLAEndtoendAutonomous2025}. Linguistic reasoning and embodied navigation are two fundamentally different forms of intelligence: the former describes the world, while the latter lives within it~\cite{harnadSymbolGroundingProblem1990, clarkBeingTherePutting1996}. Third, Density is not Biological. Dense representations are biologically implausible. Nature demonstrates that even the simplest organisms can achieve robust survival with extremely sparse neural resources~\cite{olshausenEmergenceSimplecellReceptive1996, sterlingPrinciplesNeuralDesign2017}. Their intelligence stems from the encoding of meaning by dynamic neural assemblies, not the pixel-level replication of the world~\cite{hebbOrganizationBehaviorNeuropsychological1949, buzsakiNeuralSyntaxCell2010}.

we propose the Mental Bayesian Causal World Model (MBCWM) and instantiate it as Tokenized Intent World Model (TIWM), a cognitive computing architecture. An agent should not reconstruct what it sees, but understand what it needs. As Jakob von Uexküll revealed in 1909, organisms do not inhabit an objective environment (\textit{Umgebung}) , but dwell in a self-world (\textit{Umwelt})  constructed by task significance---the world of a tick consists only of temperature, butyric acid, and hair; all else is void~\cite{uexkuellUmweltUndInnenwelt1909}. Human driving is similar: we do not reconstruct the pixels of the entire street, but focus only on affordances relevant to our current intent—such as cutting-in vehicles, intersections requiring yielding, or blind spots. MBCWM redefines intelligence as an active process of understanding through Belief-Intent Co-Evolution, rather than a mere passive response to sensory observations.

The main contributions of this work are as follows: 

(1) We propose the MBCWM and instantiate it as TIWM, a differentiable end-to-end cognitive computing architecture grounded in Belief-Intent Co-Evolution. By unifying active perception~\cite{bajcsyActivePerception1988} and intent-driven planning~\cite{fristonActiveInferenceLearning2016} through the dynamic evolution of internal representations~\cite{grushEmulationTheoryRepresentation2004}, our approach operationalizes classical cognitive robotics theories~\cite{ziemkeAreRobotsEmbodied2001, pfeiferHowBodyShapes2006, ziemkeBodyLanguageMind2007} into a scalable neural framework.

(2) We identify and validate a ``Cognitive Consistency~\cite{festigerTheoryCognitiveDissonance1957}" learning mechanism. Through Implicit Computational Replay, the system achieves a self-organizing equilibrium between belief (state understanding) and intent (future prediction)~\cite{rieglerWhenCognitiveSystem2002}, ensuring the semantic alignment of internal representations with physical affordances~\cite{gibsonTheoryAffordances19792014}.

(3) We validated the feasibility of TIWM on the nuPlan benchmark~\cite{karnchanachariLearningbasedPlanningNuPlan2024a}. In open-loop testing, we confirmed that the Belief-Intent Co-Evolution mechanism significantly enhances planning performance. In closed-loop simulations, we demonstrated the architecture’s capacity for mastering the physical world, showcasing the emergence of cognitive intelligence that transcends standard imitation learning.

\section{Related Work}

Traditional autonomous driving systems typically adopt modular pipelines that decompose the task into independent stages: perception, prediction, and planning~\cite{liBEVFormerLearningBirdsEyeView2022, shiMTRMultiAgentMotion2024, fanBaiduApolloEM2018b}. While offering interpretability, these cascaded architectures are often susceptible to error propagation and cumulative latency. Furthermore, rule-based planners frequently struggle to generalize within complex urban environments involving vulnerable road users, often resulting in rigid, non-anthropomorphic driving behaviors.
To address these limitations, modern end-to-end approaches utilize deep neural networks to map sensory inputs directly to control signals or trajectories, enabling joint optimization across the entire decision-making stack~\cite{huPlanningOrientedAutonomousDriving2023a, jiangVADVectorizedScene2023b, liNavigationGuidedSparseScene2024}. This paradigm shift has significantly improved driving smoothness and interaction capabilities.

Vision-based end-to-end approaches, such as UniAD~\cite{huPlanningOrientedAutonomousDriving2023a} and VAD~\cite{jiangVADVectorizedScene2023b}, have achieved remarkable success through multi-task learning and dense feature fusion. Nevertheless, their heavy reliance on dense representations and short-horizon reactive forecasting fundamentally constrains their long-term reasoning capabilities.To extend temporal reasoning, recent research has incorporated World Models~\cite{haRecurrentWorldModels2018, hafnerDreamControlLearning2019} and Vision-Language-Action models (VLA) frameworks~\cite{zhouOpenDriveVLAEndtoendAutonomous2025}. While these generative approaches enable forms of ``mental simulation", they frequently succumb to the ``Reconstruction Trap"—aiming to reconstruct the objective environment (\textit{Umgebung})~\cite{uexkuellUmweltUndInnenwelt1909} in its entirety. We contend that pixel-level image synthesis introduces significant high-entropy noise that is functionally orthogonal to the core driving task. Moreover, naive autoregressive concatenation of vision, language, and action~\cite{cenRynnVLA002UnifiedVisionLanguageAction2025} often leads to gradient conflicts arising from misaligned objectives~\cite{linControllableParetoMultiTask2021}. In contrast to these approaches, TIWM suggests that effective planning may not strictly require dense, objective world reconstruction. Instead, it indicates that robust behavior can emerge from an internal cognitive world (\textit{Umwelt})~\cite{uexkuellUmweltUndInnenwelt1909} that is sparse yet semantically rich.

Recent efforts have attempted to mitigate computational redundancy through sparse representations. SSR~\cite{liNavigationGuidedSparseScene2024} employs sparse tokens for planning, yet utilizes trajectory momentum primarily to assist in reconstructing future BEV features. We argue that this reverses the priority: trajectory optimization should be the ultimate objective, rather than an auxiliary task for reconstruction. Furthermore, SSR inherently assumes Markovian transitions, overlooking the non-Markovian causal evolution essential for long-term consistency. While self-supervised future reconstruction can enhance representation learning, the introduction of conflicting multi-objective losses often impedes gradient consistency~\cite{linControllableParetoMultiTask2021}, thereby hindering global optimality. Moreover, explicitly managing the weighting and frequency for multi-frame self-supervision presents a significant challenge, as it is difficult to guarantee that selected frames contain critical information. Our experimental results indicate that such approach is sub-optimal compared to a task-driven, temporally abstract semantic alignment strategy.

MotionLM~\cite{seffMotionLMMultiAgentMotion2023} and its variants~\cite{jiaAMPAutoregressiveMotion2024} reframe motion forecasting as a language modeling task, encoding multi-agent trajectories into discrete tokens. While autoregressive mechanisms circumvent Markovian limitations, these approaches necessitate intricate vectorization pre-processing and post-processing. Fundamentally, such methods model the statistical correlations of the dataset rather than the causal inference of the physical world, occasionally leading to causal inversion during complex interactive reasoning~\cite{dehaanCausalConfusionImitation2019}. Additionally, vector-based approaches lack the intrinsic spatial grounding and implicit multi-agent representation found in rasterized frameworks~\cite{gaoVectorNetEncodingHD2020a}. Furthermore, the reliance on explicit vectorized representations often introduces significant challenges in managing the combinatorial complexity of multi-agent interactions, particularly regarding the selection of interaction targets and the balancing of loss weights. Such methods inherently lack a unified multi-agent representational substrate~\cite{seffMotionLMMultiAgentMotion2023, zhengDiffusionBasedPlanningAutonomous2024}. Consequently, these approaches are frequently constrained to utilize massive data augmentation and reinforcement learning fine-tuning to compensate for these inherent limitations and achieve spatial generalization~\cite{zhengDiffusionBasedPlanningAutonomous2024, pengImprovingAgentBehaviors2025}.

Early Artificial Intelligence (AI) theories extensively explored cognitive principles in embodied systems, encompassing selective perception~\cite{bajcsyActivePerception1988, ballardAnimateVision1991, simonsGorillasOurMidst1999}, active intent~\cite{ballardDeicticCodesEmbodiment1997, raoPredictiveCodingVisual1999, fristonActiveInferenceLearning2016}, minimal representation~\cite{brooksIntelligenceRepresentation1991, grushEmulationTheoryRepresentation2004}, and interactionist computation~\cite{agreComputationalResearchInteraction1995}. A central theme was the construction of a subjective internal world, or Umwelt~\cite{brooksIntelligenceRepresentation1991, clarkBeingTherePutting1996, scheierUnderstandingIntelligence2001}. However, constrained by the computational limitations of the era, these foundational concepts remained largely theoretical. In engineering practice, implementation attempts were often restricted to loose analogies within rule-based architectures~\cite{lairdCognitiveRoboticsUsing2012, moulin-frierDACh3ProactiveRobot2017, divakarlaCognitiveAdvancedDriver2019}, lacking end-to-end differentiable realizations.

Recent trends have gravitated towards utilizing LLMs as cognitive reasoning engines~\cite{loLLMbasedRobotPersonality2025}. Yet, as statistical repositories of linguistic patterns, LLMs inherently lack physical embodiment and environmental grounding~\cite{harnadSymbolGroundingProblem1990, clarkBeingTherePutting1996}. While LeCun's World Model architecture~\cite{lecunPathAutonomousMachine2022} correctly eschews pixel-level reconstruction, its primary objective remains the pre-training extraction of abstract, high-dimensional physical laws. Furthermore, its reliance on modular components—such as configurators and critics—falls short of achieving a truly unified, organic neural system.

In contrast, we construct an internal cognitive world for the agent as illustrated in Fig.~\ref{fig:worldmodel}. Our approach operationalizes sparse coding~\cite{olshausenEmergenceSimplecellReceptive1996} through selective representations and instantiates Global Workspace~\cite{baarsGlobalWorkspaceTheory2005} dynamics via sparse token competition~\cite{grushEmulationTheoryRepresentation2004}. By orchestrating a Belief-Intent Co-Evolution, the system simultaneously distills internal representations and drives the forward evolution of the internal state. This mechanism enables the system to achieve mental simulation~\cite{decetyPowerSimulationImagining2006} within a unified end-to-end framework.
We establish Cognitive Consistency~\cite{festigerTheoryCognitiveDissonance1957} as the fundamental driver of intelligence. Specifically, autoregressive intent dynamics facilitate forward-looking, goal-oriented predictive reasoning~\cite{ballardDeicticCodesEmbodiment1997, raoPredictiveCodingVisual1999, fristonFreeenergyPrincipleUnified2010} , while task-driven backpropagation enables the retroactive modulation of perceptual targets~\cite{bajcsyActivePerception1988, denilLearningWhereAttend2012, parkEmotionalArousalEnhances2025}. This architecture effectively realizes Interactionist Artificial Intelligence~\cite{agreComputationalResearchInteraction1995}. By dynamically aligning sparse intent tokens with physical affordances, TIWM achieves planning grounded in internal active understanding rather than passive, reactive reflexes.

\section{Method}

\begin{figure*}[ht]
\vspace{-10pt}
\centering
\includegraphics[width=6.8in]{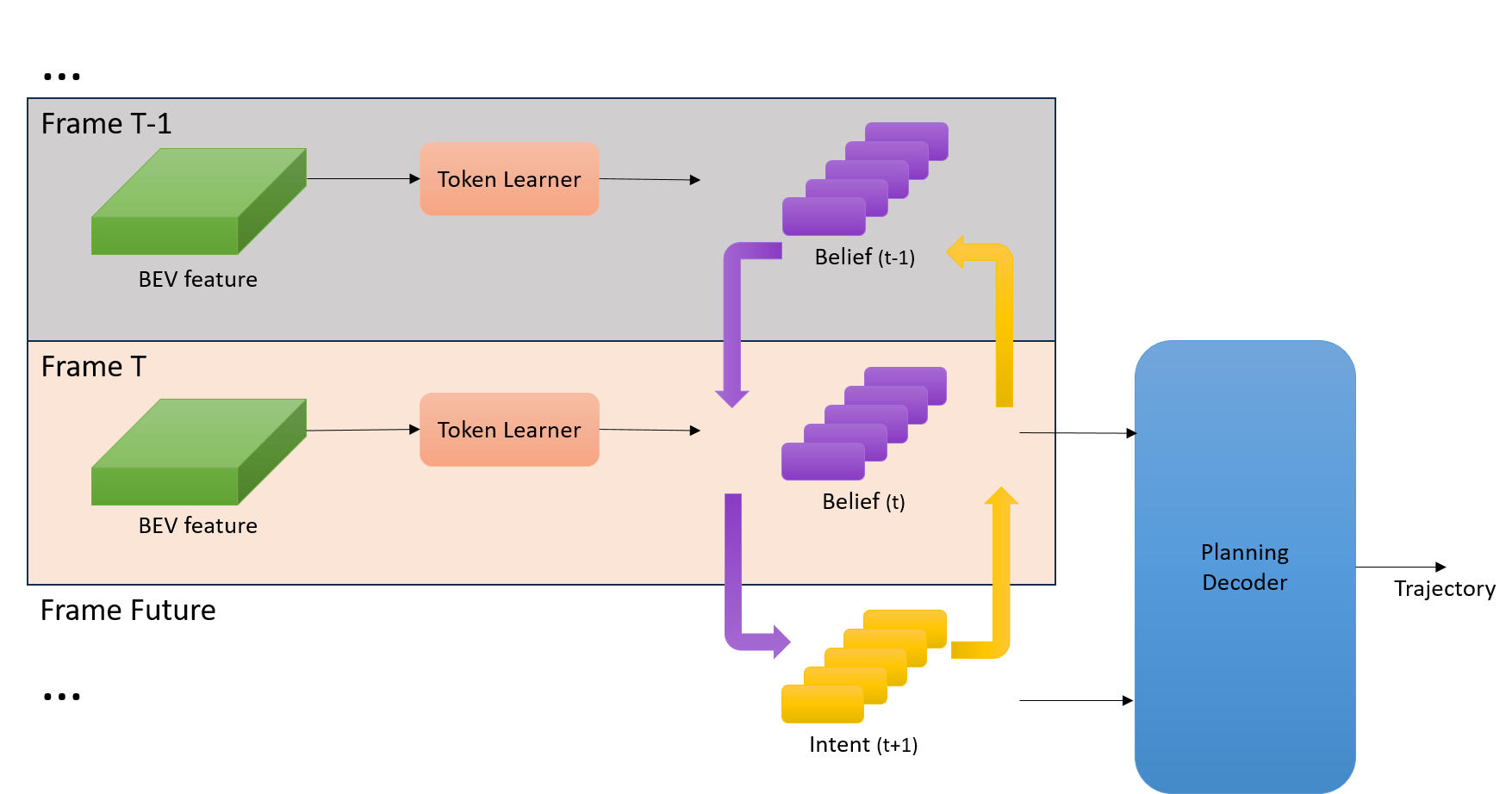}
\caption{\small \textbf{Tokenized Intent World Model: From Perception to Cognitive World.} At each timestep, sparse tokens are extracted from BEV perception features, distilling the scene into its essential semantic constituents. The architecture then autoregressively predicts future intent tokens, which serve as temporally abstract and task-oriented semantic projections of the agent's future states. The planning decoder performs joint reasoning over current sparse tokens and predicted intents to generate executable trajectories. Each intent token serves a three-fold functional role:
(1) \textbf{Representation} — compressing the core cognitive state of the environment; 
(2) \textbf{Intent} — unfolding plausible futures via autoregressive evolution;
(3) \textbf{Decision} — directly orienting the system toward goal-directed action. In this framework, the world is treated as an actionable cognitive entity—interpreted, imagined, and utilized—rather than a dense, objective reconstruction of the environment.}
\label{fig:tiwm}
\vspace{-20pt}
\end{figure*}

\subsection{Mental Bayesian Causal World Model}
Drawing inspiration from cognitive science, we establish the MBCWM as the theoretical bedrock of TIWM. The MBCWM conceptualizes the agent's internal world as a causal universe governed by the latent dynamics of spatiotemporal states.

\begin{itemize}
    \item \textbf{Belief ($T_t$)}: A state-based understanding of the environmental context, represented by a set of sparse semantic tokens.
    \item \textbf{Intent ($I_{t+1}$)}: A causal expectation of the future cognitive state, existing as a predicted token set.
\end{itemize}
Formally, the MBCWM is defined by the following four components:

\begin{align}
    T_t &= \mathcal{L}(BEV_t) \\
    I_{t+1} &= \mathcal{P}(T_{t-k:t}) \\
    \tau_{t+1:t+H} &= \mathcal{D}(I_{t+1}) \\
    \nabla_{\Theta_\mathcal{L}} \mathcal{L}_{traj} &\propto \frac{\partial \tau}{\partial I_{t+1}} \cdot \frac{\partial I_{t+1}}{\partial T_{t-k:t}} \cdot \frac{\partial T_{t-k:t}}{\partial \Theta_\mathcal{L}}
\end{align}

In this formulation, $\mathcal{L}$ denotes the Token Learner, which extracts task-relevant sparse tokens from BEV features; $\mathcal{P}$ denotes the Intent Predictor, which autoregressively generates future intent states; $\mathcal{D}$ denotes the Planning Decoder, which fuses current beliefs and future intents to decode the trajectory; Finally $\mathcal{L}_{traj}$ is the trajectory loss. Its gradient dynamically adjusts the perceptual attention mechanism via backpropagation, realizing the retroactive guidance of belief by intent.

Within the Bayesian model, history world state constitute the belief through sparse spatiotemporal representations. By projecting future intent states from historical beliefs, the neural architecture operationalizes the Mental Models (MM)~\cite{johnson-lairdMentalModelsCognitive1983}) psychological causal model characterized by symbolic deduction. Concurrently, future intents are constrained by task-specific loss functions. Through gradient backpropagation, the system achieves retroactive tuning and pruning of the belief state, effectively materializing the Force Composition (FC)~\cite{talmyForceDynamicsLanguage1988} causal model. In this process, the gradient flow acts as a ``Causal Corrective Force" that reshapes internal representations. Consequently, internal state transitions are driven by implicit dynamics composed of both forward prediction and backward propagation, thereby instantiating the Causal Models (CM)~\cite{pearlCausalityModelsReasoning2000}) of probabilistic induction.

Belief and intent \cite{bratmanIntentionPlansPractical1987a} are bidirectionally coupled through the reciprocal dynamics of autoregressive prediction and backpropagation. Guided by the implicit gradient flow, each token group undergoes a progressive transition toward a functional duality, wherein belief and intent attributes become inextricably intertwined into a single representational substrate. This dynamic co-evolution facilitates a robust convergence toward a state of Cognitive Consistency \cite{festigerTheoryCognitiveDissonance1957}, establishing a training paradigm we define as \textbf{Cognitive Consistency Learning}. By progressively embedding intentional semantic components into these representations to harmonize perception with task-oriented objectives, we designate this emergent functional unit as the \textbf{Intent Token} and formally introduce the Tokenized Intent World Model (TIWM).

\subsection{Tokenized Intent World Model}

TIWM fundamentally departs from traditional world models that prioritize the simulation of an objective environment. Instead, it constructs a subjective \textit{Umwelt} through sparse semantic tokens (Fig.~\ref{fig:tiwm}). The entire process constitutes an embodied cognitive closed-loop encompassing perception for functional signal extraction, intent for causal deduction, action for trajectory generation, and feedback for attention adjustment. In this architecture, sparse intent tokens function as neural assemblies~\cite{buzsakiNeuralSyntaxCell2010} that bridge the gap between perceptual representation and navigational execution. Through Cognitive Consistency Learning, the model aligns its internal attentional dynamics with physical affordances, ultimately achieving high-performance, reconstruction-free task planning.

\subsection{Sparse Intent Representation}
TIWM leverages upstream perception inputs to construct BEV representations, simulating a mature visual workspace to focus on decision-level behavior~\cite{ballardDeicticCodesEmbodiment1997a, vanrullenPerceptualCycles2016}. The input comprises historical and current dynamic object layers along with current map layer, including \texttt{LANE}, \texttt{INTERSECTION}, \texttt{STOP\_LINE}, \texttt{CROSSWALK}, and current baseline layer. This structured input allows the network to implicitly encode relative dynamics akin to optical flow, enabling the ego-vehicle to perceive the world's flux relative to itself and implicitly model game-theoretic interactions.

The core of our architecture is the Sparse Token Learner, which compresses the high-dimensional BEV tensor $\mathbf{X} \in \mathbb{R}^{C\times H\times W}$ into a compact set of $N{=}16$ semantic tokens $\mathbf{T} \in \mathbb{R}^{N\times D}$:
\begin{equation}
\mathbf{T} = \sum_{hw} \text{Softmax}(\phi(\mathbf{X})) \odot \mathbf{X},
\end{equation}
The spatial attention weights are computed via a lightweight convolutional network:
\begin{equation}
\phi(\mathbf{X}) = \text{Conv}_2(\text{ReLU}(\text{Conv}_1(\mathbf{X}))).
\end{equation}
This mechanism condenses the $224{\times}224$ grid into 16 global workspace units~\cite{guanWorldModelsAutonomous2024}, operationalizing sparse coding~\cite{olshausenEmergenceSimplecellReceptive1996} to capture the most decision-relevant information.

\subsection{Intent Chain Evolution}
TIWM conceptualizes planning as a dynamic interplay between belief (state
understanding) and intent (future prediction). Rather than reconstructing dense future states, the model autoregressively evolves a sparse Intent Chain from historical token sequences. Given four historical token sets $\mathbf{T}_{t-3:t}$, a learnable future query $\mathbf{Q} \in \mathbb{R}^{N \times D}$, and a causal attention mask $\mathcal{M}$, we compute the future intent $\mathbf{I}_{t+1}$ as follows:
\begin{equation}
\begin{split}
[\mathbf{T}_{t-3:t}, \mathbf{I}_{t+1}] = &\text{TransformerEncoder}\\
&(\text{Concat}(\mathbf{T}_{t-3:t}, \mathbf{Q}); \mathcal{M}).
\end{split}
\end{equation}
The causal mask $\mathcal{M}$ ensures that future intent tokens only attend to historical and current information, precluding temporal leakage. This architecture establishes the mechanism of Belief-Intent Co-Evolution, where the forward pass synthesizes intent from belief, while intent-based gradients implicitly steer the belief encoder to focus on task-relevant affordances. This bidirectional flow endows tokens with a functional duality, functioning simultaneously as grounded representations of the present and causal seeds for the future.

\subsection{Multi-Modal Motion}
TIWM conceptualizes planning as the action trajectory that bridges the current belief state to the projected future intent. The executable trajectory is derived as follows:
\begin{equation}
\hat{\tau} = \phi(\mathbf{T}{t}, \mathbf{I}{t+1}) \end{equation}
where $\hat{\tau}$ denotes the predicted 8-second trajectory.


The decoder maintains 16 learnable motion primitives. To enforce sparse activation and soft competition, these primitives are fused into a dynamic representation using a weight vector $w = \text{Softmax}(\Phi(\mathbf{I}_{t+1}))$, where $\Phi$ is a learnable projection. This gating mechanism selects the most physically consistent motion candidates from the primitive bank. The resulting representation subsequently interacts with current beliefs $\mathbf{B}_t$ and intents $\mathbf{I}_{t+1}$ through a cross-attention mechanism for spatiotemporal refinement. Finally, a multi-layer perceptron head decodes the synthesized features into an 80-point executable trajectory.

\subsection{Task-Driven Semantic Alignment}
The training objective of TIWM is intentionally minimalist, relying exclusively on the ultimate planning outcome:

\begin{equation}
\mathcal{L}_{\text{total}} = \mathcal{L}_{\text{traj}} = \lambda_{\text{traj}} \cdot \text{SmoothL1}(\hat{\tau}, \tau_{gt}) \label{eq:loss_traj}
\end{equation}
where $\tau_{gt}$ represents the ground-truth 8-second trajectory.

For ablation studies, we also evaluate an auxiliary reconstruction loss to align predicted intents with ground-truth tokens:
\begin{equation}
\begin{split}
\mathcal{L}_{\text{intent}} = \lambda_{\text{intent}} &\| \text{normalize}(\mathbf{I}_{t+1}) \\
&- \text{normalize}(\mathbf{T}_{t+1}^{\text{gt}}) \|_2,
\end{split}
\end{equation}
However, our experimental evidence confirms that while a joint loss $\mathcal{L}_{\text{total}} = \mathcal{L}_{\text{traj}} + \mathcal{L}_{\text{intent}}$ can be formulated, such explicit supervision results in performance degradation. This finding suggests that the Intent Token should emerge as a task-driven functional unit, rather than requiring additional reconstruction supervision of future states.

\section{Experiments}

\subsection{Experimental Setup}

The experimental evaluation is structured into two primary components: open-loop evaluation, aimed at validating trajectory prediction performance and the efficacy of individual architectural components; and closed-loop simulation, which focuses on verifying computational feasibility, performance within finite task scenarios, and the emergence of human-like cognitive driving behaviors.

\textbf{Dataset and Metrics.} Open-loop experiments were conducted using the nuPlan autonomous driving dataset~\cite{karnchanachariLearningbasedPlanningNuPlan2024a}. Specifically, we extracted 360 scenarios from the Mini Split, sampling at 2Hz (0.5s per frame) to construct a dataset comprising 13k frames. This data was partitioned into training and validation sets at a 9:1 ratio to conduct ablation studies and assess performance upper bounds under high distributional similarity. To investigate scaling properties, we randomly selected 10,000 scenarios from the Boston full dataset, maintaining an 8.5:1.5 scenario-based split. These experiments encompassed five training subsets of varying magnitudes (10k, 50k, 100k, 200k, and 300k frames), with evaluation performed on a fixed 54k frames validation set. Performance is quantified using Average Displacement Error (ADE) and Final Displacement Error (FDE) over an 8-second predicted horizon consisting of 80 trajectory points.

\textbf{Implementation Details.} TIWM utilizes a ResNet-18 backbone to encode BEV inputs generated by the RasterFeatureBuilder. While these rasters provide necessary environmental context such as lane markings and road boundaries, we strictly exclude ego status, vectorized reference lines, and structured HD map priors. This constraint necessitates that the model performs autonomous reasoning and path synthesis from raw spatiotemporal features, rather than following pre-defined reference paths. The model processes 4 historical frames (2s) as input to predict an 80-point future trajectory over an 8-second horizon. All experiments were conducted using the AdamW optimizer on a single NVIDIA RTX 5090 GPU.

\textbf{Closed-loop Simulation Details.} Following the open-loop setup, closed-loop evaluation is performed using models trained for 500 epochs on the same 360 Mini Split scenarios. This training set remains strictly disjoint from the val14 validation set \cite{daunerPartingMisconceptionsLearningbased2023}. To ensure closed-loop stability and mitigate distribution shift, we adopt a three-stage curriculum learning strategy, progressively transitioning from foundational imitation to augmented state refinement and finally to cognitive convergence. Post-processing is limited to a median method for heading determination and outlier filtering to maintain the stability of the nuPlan LQR controller.

\begin{figure*}[t]
\vspace{-10pt}
  \centering
  \begin{minipage}[c]{0.65\textwidth}
    \centering
    \captionof{table}{\small \textbf{Ablation Analysis of TIWM Components.} We evaluate the contributions of Active Perception, Multi-Modal Planning, and Belief–Intent Co-Evolution.}
    \label{tab:ablation}
    \small
    \setlength{\tabcolsep}{3.5pt} 
    \begin{tabular}{c|cccc|cc}
      \toprule
      \multirow{2}{*}{\textbf{ID}} & \textbf{Active} & \textbf{Multi-Modal} & \textbf{Belief–Intent} & \textbf{Supervision} & \multirow{2}{*}{\textbf{ADE (m)}} & \multirow{2}{*}{\textbf{FDE (m)}} \\
      & \textbf{Perception} & \textbf{Planning} & \textbf{Co-Evolution} & \textbf{Strategy} & & \\
      \midrule
      Ref. & \multicolumn{4}{c|}{Official Pretrained ResNet-50 RasterModel} & 1.149 & 2.444 \\ 
      \midrule
      1 & \xmark & \xmark & \xmark & N/A & 0.881 & 2.047 \\
      2 & \cmark & \xmark & \xmark & N/A & 0.572 & 1.152 \\
      3 & \cmark & \cmark & \xmark & N/A & 0.551 & 1.192 \\
      4 & \cmark & \cmark & \xmark & Sparse Recon. & 0.568 & 1.209 \\
      5 & \cmark & \cmark & \cmark & Sparse Recon. & 1.029 & 2.625 \\
      6 & \cmark & \cmark & \cmark & Semantic Align. & \underline{0.388} & \underline{0.858} \\
      7 & \cmark & \xmark & \cmark & Semantic Align. & \textbf{0.378} & \textbf{0.795} \\
      \bottomrule
    \end{tabular}
  \end{minipage}
  \hfill 
  \begin{minipage}[c]{0.31\textwidth}
    \centering
    \includegraphics[width=\textwidth]{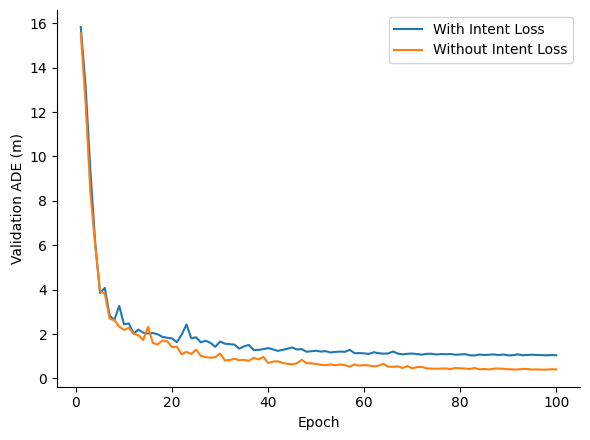}
    \caption{\small \textbf{Comparison of Supervision Strategies}}
    \label{fig:comparison}
  \end{minipage}
\vspace{-15pt}
\end{figure*}

\subsection{Ablation Study}

We conducted ablation experiments on various architectural variants, training each variant for 100 epochs on the 360-scenario dataset. As shown in Table~\ref{tab:ablation}, active perception based on sparse activation significantly enhances model performance. The integration of future intent and the implementation of the Belief-Intent Co-Evolution mechanism resulted in a further marked increase in prediction accuracy. Specifically, the ADE of the Co-Evolution mechanism decreased by 61\% compared to the baseline and by 66\% compared to the official pre-trained ResNet-50 RasterModel. Notably, our experiments demonstrate that the \textit{Sparse Recon.} strategy, which introduces explicit intent reconstruction loss as a self-supervised signal, results in lower performance compared to the \textit{Semantic Align}. strategy (Fig.~\ref{fig:comparison}). This suggests that implicit future intent alignment is superior to forced explicit reconstruction, as it allows the model to flexibly focus on critical future semantic cues. The Belief-Intent Co-Evolution mechanism achieves exceptional prediction accuracy in in-distribution scenarios, which is highly significant for enabling autonomous robotic evolution through small-sample learning.


\subsection{Convergence Study}

\begin{figure}[ht]
\vspace{-5pt}
\centering
\includegraphics[width=3.4in]{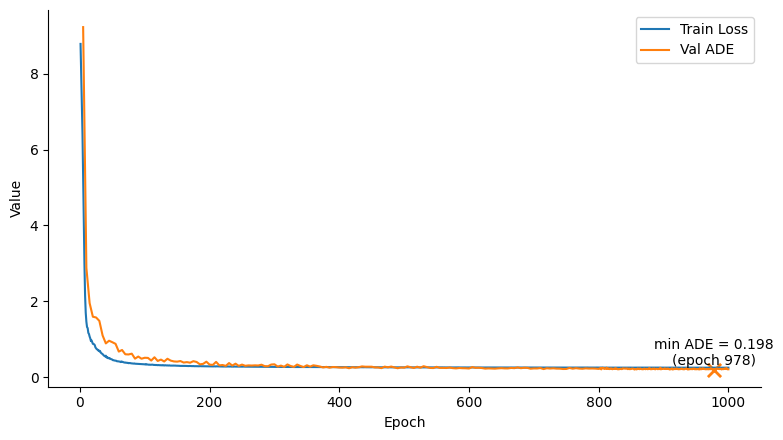}
\caption{\small \textbf{Training Loss and Validation ADE Convergence}}
\label{fig:convergence}
\vspace{-15pt}
\end{figure}

As illustrated in Fig.~\ref{fig:convergence}, the Belief-Intent Co-Evolution mechanism exhibits a sustained performance growth trend over 1,000 epochs on the 360 scenarios dataset. Even at this stage, the model has not reached full convergence, providing strong evidence for the efficacy of Cognitive Consistency Learning. During training, tokens maintain a dynamic equilibrium under the dual influence of belief and intent, allowing the internal attention mechanism to continuously align with physical affordances. Notably, our supplementary experiments reveal that data augmentation acts as a powerful catalyst for this cognitive alignment. Models trained with augmentation for only 500 epochs consistently outperform those trained for 1,000 epochs on raw data. This suggests that high-entropy augmented samples effectively accelerate the self-organized convergence between internal representations and environmental affordances.

\subsection{Scaling Study}

\begin{figure}[ht]
\vspace{-10pt}
\centering
\includegraphics[width=3.4in]{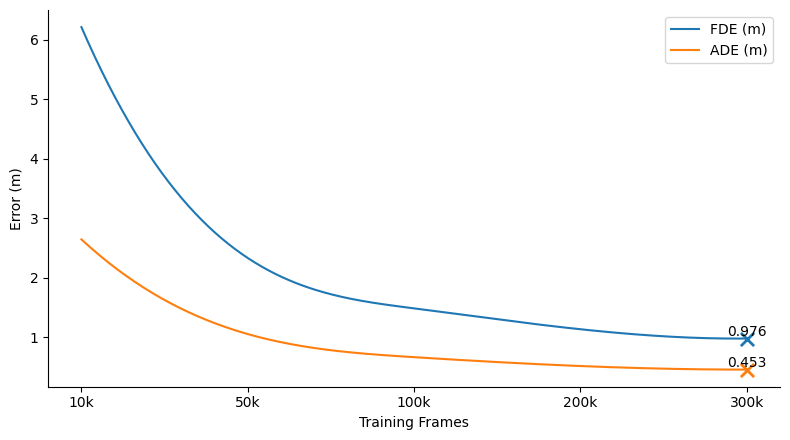}
\caption{\small \textbf{Data Scaling Capability}}
\label{fig:scaling}
\vspace{-15pt}
\end{figure}

As illustrated in Fig.~\ref{fig:scaling}, planning performance within the Boston dataset exhibits a pronounced scaling trend as data volume increases across 100 training epochs. This validates that the TIWM architecture, underpinned by the Belief-Intent Co-Evolution mechanism, effectively distills incremental data to refine its internal model of physical reality. Furthermore, these results identify the performance upper bound of the planning module when provided with high-fidelity perception inputs. This offers a critical insight for the community: using datasets like nuScenes~\cite{caesarNuScenesMultimodalDataset2020} may reveal that planning bottlenecks often stem from perceptual noise rather than algorithmic limitations. Notably, we emphasize that the inherent planning complexity of nuPlan significantly surpasses that of nuScenes.

\subsection{Attention Study}

\begin{figure*}[ht]
\vspace{-10pt}
\centering
\includegraphics[width=6.8in]{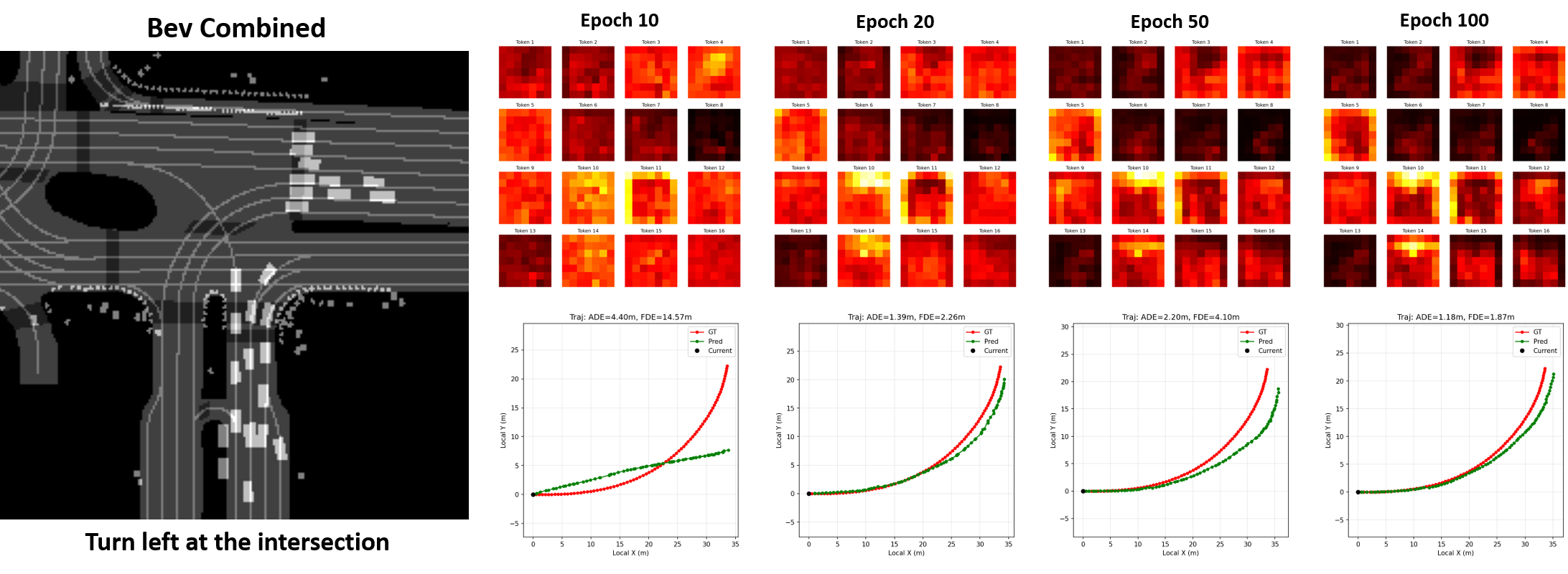}
\caption{\small \textbf{Visualization of Token Attention Maps.} This figure illustrates the synergistic refinement of attention maps and trajectory predictions across successive training epochs. As learning progresses, the attention mechanism undergoes spontaneous functional specialization, increasingly localizing on critical regions of the BEV grid. The resulting emergence of structured and hierarchical attentional patterns directly correlates with the observed refinement in trajectory planning quality.}
\label{fig:attention_map}
\vspace{-15pt}
\end{figure*}

Visualization of the attention maps (Fig.~\ref{fig:attention_map}) offers empirical substantiation for the emergence of ``Cognitive Consistency''~\cite{festigerTheoryCognitiveDissonance1957} within the TIWM architecture. While initially diffuse, the attention weights progressively self-organize into specialized functional units as training proceeds. The 16 sparse tokens demonstrate a spontaneous partitioning of the BEV space, allocating dedicated neural resources to key environmental affordances—including lane boundaries, agent dynamics, and intersection geometries. This evolution from stochastic distributions to structured, hierarchical attention confirms that the model effectively distills task-relevant invariants from the mature visual workspace, establishing a robust representational foundation for high-fidelity planning.

\subsection{Existence Proof}

To simulate the survival challenges of an embodied agent in specific environments, we designed a ``Chosen-Plaintext Attack''~\cite{goldwasserProbabilisticEncryption1984} experimental strategy. As shown in Table~\ref{tab:mini_val14}, a model trained on 360 scenarios successfully mastered the all 12 high-difficulty scenarios within the val14 validation set in Mini Split, achieving a Non-Reactive (NR) score of 88 and a Reactive (R) score of 90. This proves that within finite task scenarios, there exists a high-entropy subset that enables TIWM to achieve high-performance operation. Following the ``Biological Minimal Cost Principle"~\cite{sterlingPrinciplesNeuralDesign2017}, it is more pragmatically significant for a robotic system to achieve reliable survival in limited environments at minimal cost than to blindly pursue Artificial General Intelligence (AGI). We cite the performance returns of the representative SOTA Diffusion Planner~\cite{zhengDiffusionBasedPlanningAutonomous2024} to demonstrate that simply increasing data volume yields extremely low marginal returns for improving performance in specific tasks.

\begin{table}[t]
\vspace{5pt}
\centering
\caption{\small \textbf{nuPlan Mini Split Val14 Scenario Testing Results} 
}
\setlength{\tabcolsep}{6pt} 
\begin{tabular}{l c c c c}
\toprule
\multirow{2}{*}[-0.15ex]{\textbf{Planner}} & 
\multicolumn{2}{c}{\textbf{Training Data}} & 
\multicolumn{2}{c}{\textbf{Val14 Score}} \\ 
\cmidrule(lr){2-3} \cmidrule(lr){4-5}
& \textbf{Scenarios} & \textbf{Frames} & \textbf{NR} & \textbf{R} \\
\midrule
TIWM Planner(Ours) & 360 & 12k & 88.05 & 89.54 \\ 
\midrule
\multirow{3}{*}{Diffusion Planner~\cite{zhengDiffusionBasedPlanningAutonomous2024}} 
& 10k & 10k & 77.64 & 86.33 \\ 
& 300k & 300k & 88.94 & 94.62 \\ 
& 10M & 10M & 98.08 & 95.7 \\ 
\bottomrule
\end{tabular}
\label{tab:mini_val14}
\vspace{-15pt}
\end{table}

\subsection{Cognitive Behavior Emergence}

\begin{figure*}[ht]
\vspace{-10pt}
\centering
\begin{minipage}[b]{0.66\linewidth}
\centering
\includegraphics[width=\linewidth]{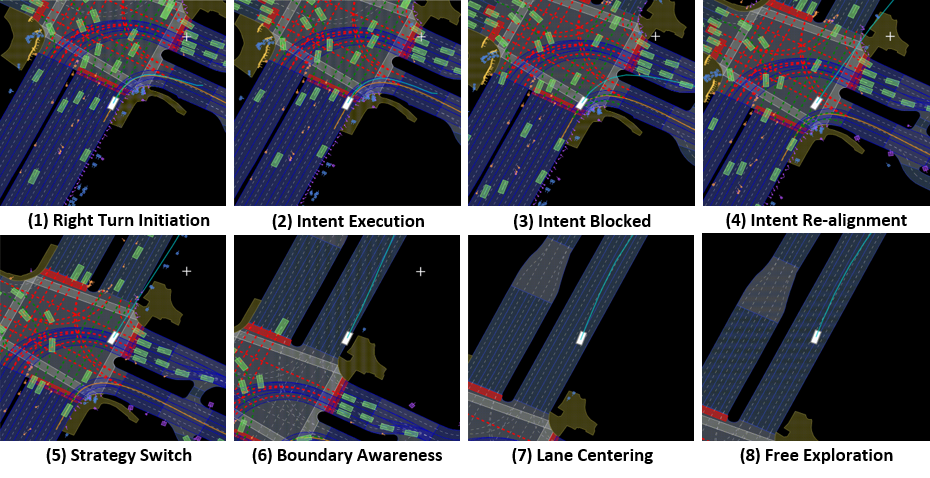}
\caption{\small \textbf{Decision Resilience and Autonomous Recovery in Intersection Scenario.} In this intersection scenario, when a planned lane change failed due to cumulative errors, the ego-vehicle avoided stalling or logical deadlock. Instead, it demonstrated decision-making resilience by autonomously adapting its strategy to safely navigate the intersection. Upon approaching the risk boundary at the road edge, the vehicle spontaneously executed a corrective maneuver to regain lane centering, subsequently transitioning into free exploration without pre-assigned tasks.}
\label{fig:cognitive_behaviour}
\end{minipage}
\hfill 
\begin{minipage}[b]{0.33\linewidth}
\centering
\includegraphics[width=\linewidth]{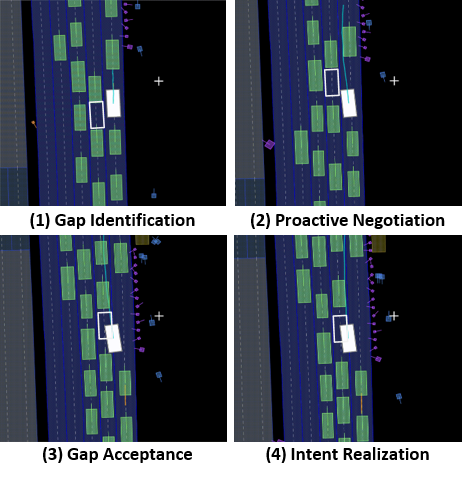}
\caption{\small \textbf{Implicit Gaming Capability in Lane Change.} In dense traffic, the ego-vehicle exhibits clear proactive interaction intent. Through autonomous negotiation with surrounding dynamic agents, it successfully completes complex cut-in maneuvers in the reactive closed-loop simulation.}
\label{fig:gaming_lane_change}
\end{minipage}
\vspace{-30pt} 
\end{figure*}

In closed-loop simulations, TIWM exhibits emergent cognitive behaviors that transcend conventional imitation learning. As illustrated in Fig.~\ref{fig:cognitive_behaviour} and Fig.~\ref{fig:gaming_lane_change}, the model spontaneously develops a sophisticated understanding of dynamic environments without explicit programming. This is manifested through robust self-recovery, autonomous exploration, and implicit gaming strategies during interaction. The emergence of these patterns provides compelling evidence that sparse intent tokens effectively capture the underlying causal structure of driving, marking a fundamental transition from statistical imitation to causal-consistent reasoning.

\subsection{Feasibility Proof}

\begin{table}[t]
\vspace{5pt}
\centering
\caption{\small \textbf{nuPlan Val14 Benchmark Results.} \textbf{*}: Using pre-searched reference lines as model input provides prior knowledge.}
\begin{tabular}{llcc}
\toprule
\multirow{2}{*}[-0.15ex]{\makecell[l]{\textbf{Type}}} & \multirow{2}{*}[-0.15ex]{\makecell[l]{\textbf{Planner}}} & \multicolumn{2}{c}{\textbf{Val14}} \\ 
\cmidrule(lr){3-4} 
&  & \textbf{NR} & \textbf{R} \\ \midrule
\textcolor{gray}{Expert} & \textcolor{gray}{Log-replay}  & \textcolor{gray}{93.53}          & \textcolor{gray}{80.32} \\ \midrule
\multirow{4}{*}[-0.15ex]{\makecell[l]{Rule-based \\ \& Hybrid}}
& IDM~\cite{treiberCongestedTrafficStates2000}                    & 75.60          & 77.33 \\ 
& PDM-Closed~\cite{daunerPartingMisconceptionsLearningbased2023}             & 92.84          & 92.12 \\
& PDM-Hybrid~\cite{daunerPartingMisconceptionsLearningbased2023}             & 92.77          & 92.11 \\ 
& GameFormer~\cite{huangGameFormerGametheoreticModeling2023}             & 79.94          & 79.78 \\ 
\midrule
\multirow{7}{*}[0.5ex]{\makecell[l]{Learning-based}}
& PDM-Open~\cite{daunerPartingMisconceptionsLearningbased2023}\textbf{*}             & 53.53          & 54.24 \\ 
& UrbanDriver~\cite{duLearningDemonstrationsCritical2023}            & 68.57          & 64.11 \\ 
& GameFormer~\cite{huangGameFormerGametheoreticModeling2023} w/o refine.        & 13.32          & 8.69  \\ 
& PlanTF~\cite{chengRethinkingImitationbasedPlanners2024}                 & 84.27          & 76.95 \\ 
& PLUTO~\cite{chengPLUTOPushingLimit2024} w/o refine.\textbf{*}            & 88.89         & 78.11 \\ 
& Diffusion Planner~\cite{zhengDiffusionBasedPlanningAutonomous2024}       & 89.87         & 82.80 \\  
& \textbf{TIWM Planner(Ours)}        & 30.46         & 34.64 \\  
\bottomrule
\end{tabular}
\label{tab:val14}
\vspace{-15pt}
\end{table}

We evaluated the model on the full val14 validation set. As shown in Table~\ref{tab:val14}, TIWM achieved a NR score of 30 and a R score of 35, using only 360 scenarios (1.5 hours of driving data) and 2.5 hours of training. These results demonstrate that TIWM, with only 12.45M parameters, effectively decodes physical world affordances through a sparse raster-based architecture without any ego status input, validating the computational feasibility of the framework. Notably, in terms of planning scores, TIWM outperformed GameFormer~\cite{huangGameFormerGametheoreticModeling2023}, which utilizes explicit game-theoretic modeling.

\section{Discussion}

The development of MBCWM represents an epistemological shift from objective environmental modeling toward the construction of a functional internal world. The TIWM architecture serves as a computational realization of von Uexküll’s insight: intelligence does not merely inhabit an objective physical environment (\textit{Umgebung})  but operates within a subjective world of meaning (\textit{Umwelt})~\cite{uexkuellUmweltUndInnenwelt1909}. Unlike traditional world models focused on pixel-level reconstruction, TIWM constructs a sparse \textit{Umwelt} composed of functional signs (\textit{Merkmal})~\cite{uexkuellUmweltUndInnenwelt1909}. Biologically, this is analogous to a tick's specialized sensitivity to butyric acid; computationally, it manifests as the extraction of causal affordances. These sparse tokens form the semiotic substrate for causal reasoning, empowering the system to replace passive, reactive reflexes with proactive internal understanding and foresightful imagination.

Within this framework, we identify Cognitive Consistency~\cite{festigerTheoryCognitiveDissonance1957} as the core driver of learning. Through prolonged training, the system spontaneously converges toward a dynamic equilibrium where internal belief states and intent states achieve high semantic alignment. This process dynamically filters scene fragments most relevant to intent generation, gradually aligning internal representations with the affordance structure of the physical world. This equilibrium mirrors biological phenomena such as Neural Replay~\cite{huangReplaytriggeredBrainwideActivation2024}, where the model refines its internal world through ``Computational Replay". Based on this, we propose two fundamental hypotheses for embodied AI: (H1) the efficacy of embodied intelligence depends less on parameter scale or data volume than on the degree of semantic alignment between internal attentional dynamics and physical affordances; (H2) the Intent Token serves as a functional neural computational unit, analogous to the biological neural assembly~\cite{hebbOrganizationBehaviorNeuropsychological1949, buzsakiNeuralSyntaxCell2010}, rather than a mere scalar neuron.

This work further confirms the feasibility of small-sample autonomous evolution and sparse token self-organization, validating the engineering value of the neural replay mechanism. By acquiring basic driving competence through Belief-Intent Co-Evolution using limited samples, our results suggest that high-entropy samples play a pivotal role in the performance scaling of autonomous robotic system. This opens a new path for the autonomous evolution of pre-trained cognitive robots: transitioning from data-driven to cognitive-driven learning via high-quality computational replay and the reactivation of attention weights.

\textbf{Limitations and Open Questions.} 
Despite promising results, the stability of closed-loop performance and its correlation with data distribution, attention collapse, and non-Markovian evolution require further quantitative study. Furthermore, closed-loop simulations suggest that data augmentation can offset cumulative errors. This implies that once predictive precision surpasses a critical threshold, the ``Open-loop/Closed-loop Paradox"~\cite{liEgoStatusAll2024} may be resolved through the integration of reinforcement learning. Future work should explore higher-order trajectory representations and advanced decoder architectures to validate TIWM in large-scale, real-world scenarios.

\section{Conclusion}

We have proposed the MBCWM theory and implemented the TIWM, a novel cognitive computational architecture. Its core principle posits that intelligence stems not from pixel-perfect fidelity, but from the Cognitive Consistency between the agent's internal cognitive world and physical reality. On the nuPlan benchmark, open-loop validation confirms that the Belief-Intent Co-Evolution mechanism effectively enhances planning performance. In closed-loop simulations, we verified the computational feasibility of our approach, showcasing cognitive intelligence that transcends simple statistical imitation. We identify Cognitive Consistency Learning as the primary mechanism: through implicit computational replay, the system achieves a self-organized equilibrium between belief and intent, continuously improving performance on finite data. Our neuro-symbolic fusion architecture illuminates a path for robots to develop through internal understanding and imagination.
In future work, we will: (1) extend TIWM to large-scale closed-loop testing and real-vehicle deployment, broadening its application to multi-agent systems and conversational agents; (2) utilize computational replay mechanisms within cognitive architectures to enable autonomous survival and evolution of small-sample pre-trained cognitive robots in novel environments.



\bibliographystyle{plainnat}
\bibliography{references}

\end{document}